\documentclass[letterpaper, 10 pt, conference]{ieeeconf}  % Comment this line out if you need a4paper

\IEEEoverridecommandlockouts                              % This command is only needed if 
\usepackage{amsmath} % assumes amsmath package installed
\usepackage{romannum}
\usepackage{graphicx}
\usepackage{amssymb}
\usepackage{graphics} % for pdf, bitmapped graphics files
\usepackage{epsfig} % for postscript graphics files
\usepackage{mathptmx} % assumes new font selection scheme installed
\usepackage{times} % assumes new font selection scheme installed
\usepackage{amsmath} % assumes amsmath package installed
\usepackage{amssymb}  % assumes amsmath package installed
\usepackage{array}
\usepackage{mathtools}
\usepackage{cite}
\usepackage{siunitx}
\usepackage{array}% http://ctan.org/pkg/array
\usepackage{units}% http://ctan.org/pkg/array
\usepackage{booktabs}
\usepackage{nomencl}
\usepackage{tabu}
\usepackage{romannum}
\usepackage{multirow}
\usepackage{subcaption}
\usepackage{graphicx}% http://ctan.org/pkg/graphicx
\usepackage{romannum}
\usepackage{stackengine}
\usepackage{url}
\usepackage{amsmath}
\usepackage{algorithm}
\usepackage[noend]{algpseudocode}

\title{\LARGE \bf
Game-Theoretic Model Predictive Control with Data-Driven Identification of Vehicle Model for Head-to-Head Autonomous Racing
}

\author{Chanyoung Jung$^{1}$, Seungwook Lee$^{1}$, Hyunki Seong$^{1}$, Andrea Finazzi$^{2}$ and David Hyunchul Shim$^{1}$%
\thanks{$^{1}$Chanyoung Jung, Seungwook Lee, Hyunki Seong, and David Hyunchul Shim are with the School of Electrical Engineering, Korea Advanced Institute of Science and Technology, Daejeon, South Korea. 
{\tt\small cy.jung@kaist.ac.kr} ,
{\tt\small seungwook1024@kaist.ac.kr},
{\tt\small hynkis@kaist.ac.kr},
{\tt\small hcshim@kaist.ac.kr}}%
\thanks{$^{2}$Andrea Finazzi is with the KAIST Robotics Program, Korea Advanced Institute of Science and Technology, Daejeon, South Korea.
        {\tt\small finazzi@kaist.ac.kr}}%
\thanks{\emph{(Chanyoung Jung, Seungwook Lee and Hyunki Seong contributed equally to this work)}}%
}

\begin{document}

\maketitle
\thispagestyle{empty}
\pagestyle{empty}

%%%%%%%%%%%%%%%%%%%%%%%%%%%%%%%%%%%%%%%%%%%%%%%%%%%%%%%%%%%%%%%%%%%%%%%%%%%%%%%%
\begin{abstract}
% Head-to-head autonomous racing, which is a type of autonomous racing, is driven by the limits of handling; however, head-to-head autonomous racing, presents technical difficulties since more than one participant participate in this type of racing. 
% Head-to-head autonomous racing은 최근 
% Edge-case의 자율 주행 기술 개발을 위해 head-to-head autonomous racing은 최근 산업계 및 학계에서 많은 관심을 받고 있는 연구 topic이다. 
Resolving edge-cases in autonomous driving, head-to-head autonomous racing is getting a lot of attention from the industry and academia. In this study, we propose a game-theoretic model predictive control (MPC) approach for head-to-head autonomous racing and data-driven model identification method. For the practical estimation of nonlinear model parameters, we adopted the hyperband algorithm, which is used for neural model training in machine learning. The proposed controller comprises three modules: 1) game-based opponents' trajectory predictor, 2) high-level race strategy planner, and 3) MPC-based low-level controller. The game-based predictor was designed to predict the future trajectories of competitors. Based on the prediction results, the high-level race strategy planner plans several behaviors to respond to various race circumstances. Finally, the MPC-based controller computes the optimal control commands to follow the trajectories. The proposed approach was validated under various racing circumstances in an official simulator of the Indy Autonomous Challenge. The experimental results show that the proposed method can effectively overtake competitors, while driving through the track as quickly as possible without collisions.

\end{abstract}

%%%%%%%%%%%%%%%%%%%%%%%%%%%%%%%%%%%%%%%%%%%%%%%%%%%%%%%%%%%%%%%%%%%%%%%%%%%%%%%%
\section{INTRODUCTION}
Recently, prominent races related to autonomous driving have been actively organized to solve real-world problems in ``edge case'' scenarios beyond the autonomous driving technology in general situations, such as Roborace\cite{roborace}, Indy Autonomous Challenge\cite{iac}, and DARPA-RACER\cite{racer}. Autonomous racing can be divided into three categories: 1) time trial, 2) 1:1 racing, and 3) head-to-head autonomous racing, and all three types of racing are driven by limits of handling; however, head-to-head autonomous racing, where there is more than one opponent, presents more technical difficulties. In this study, we focused on head-to-head autonomous racing.

The MPC framework, which finds the optimal control commands based on the system model, while satisfying a set of constraints, is the most widely used controller for high-speed autonomous driving. 
% In \cite{kabzan2020amz}, the objective function of MPC maximized the progress along the reference path by integrating a linear model of the vehicle progression along the centerline of the track. 
In \cite{kabzan2020amz}, they formulated the objective function of MPC which aims to maximize the progress along the reference path by integrating a non-linear vehicle model for time trial racing.
% Using linear programming, they solved the optimization problem in real time.
In \cite{liniger2015optimization}, they proposed a model predictive contouring control (MPCC) approach that can follow the reference path and avoid stationary obstacles. They demonstrated the real-time capability of the proposed method through real-world experiments using a 1:43 scale vehicle.
% Along with various objective functions, the MPC performance for autonomous racing has been verified in various ways.
However, most of the existing studies have focused mainly on the MPC formulation based on the assumption that the vehicle and tire model parameters are known in advance.

Unlike time trials, head-to-head racing players should be able to predict future trajectories that reflect the intentions and strategies of other opponents and drive through the track as quickly as possible, while balancing safety and aggressiveness. 
% In \cite{buyval2017deriving}, an MPC that can overtake the other competitors was proposed by modeling the nonlinear system model of the vehicle and surrounding vehicles as a point-mass model. 
In \cite{buyval2017deriving}, an MPC that can overtake the other competitors was proposed. 
The MPC objective function incorporates the euclidean distance between the ego-vehicle and the opponents, penalizing collision-prone trajectories. 
The authors of \cite{wang2019game1} proposed a game-theoretic planner for overtaking in a two-car racing scenario. They used an iterative best response algorithm (IBR) that seeks for the Nash equilibrium within the joint trajectory of the two vehicles. Similarly, in \cite{liniger2019noncooperative}, a game-theoretic approach was proposed to model racing as a non-cooperative non-zero-sum game, while assuming open-loop information structures. Although various approaches have been suggested for autonomous racing with others, head-to-head racing remains challenging, and much work remains to be done.

\begin{figure*}[t]
    \centering
    \includegraphics[width=1\textwidth]{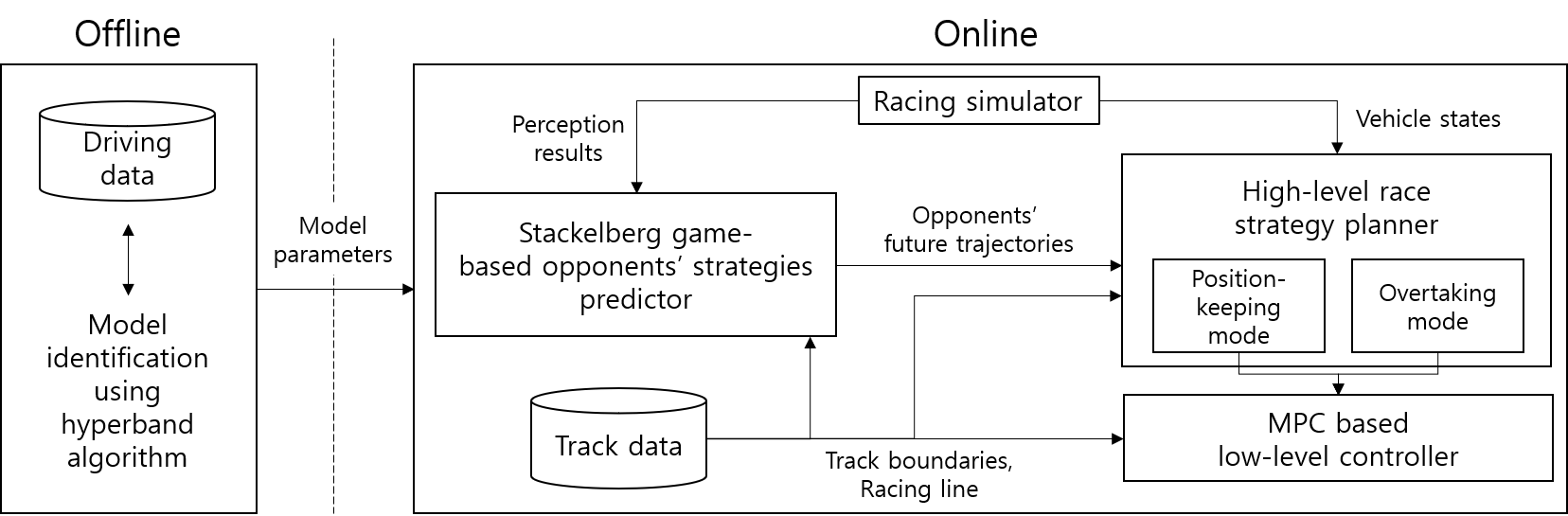}
    \caption{Overview of the proposed approach for head-to-head autonomous racing.}
    \label{fig:overview}
\end{figure*} 

% \begin{figure*}[t]
%     \centering
%     \includegraphics[width=0.99\textwidth]{figure/result_race1.png}
%     \caption{Overview of the proposed approach for head-to-head autonomous racing.}
%     \label{fig:overview}
% \end{figure*}

Herein, we present the Stackelberg game-theoretic MPC for head-to-head autonomous racing, along with a data-driven optimization-based approach to identify the system model parameters that underlie MPC. Specifically, for model identification, we applied the hyperband algorithm, which is a widely used practical approach in neural model learning. With the obtained models, we designed a race-oriented controller, which can respond to various race circumstances without collision with surrounding vehicles. Our proposed controller comprises three modules: 1) Stackelberg game-based opponents' trajectory predictor, 2) high-level race strategy planner, and 3) MPC-based low-level controller. Finally, the proposed method was verified based on a head-to-head racing scenario in a realistic vehicle simulator.

\section{Model identification using Hyperband algorithm}
% For high-performance MPC-based racing, system model parameters with high nonlinearity are required. The inaccuracy in the parameter can result in model error, which can be propagated over the horizon. Consequently, this reduces the prediction accuracy of MPC and lowers the optimality of the solution. However, finding the correct model parameters is challenging and requires many heuristic search processes. 
For MPC-based vehicle control, highly nonlinear model parameters are required. The inaccuracy in the parameters can result in model error and it is propagated over the MPC horizon. Consequently, this reduces the prediction accuracy of MPC and lowers the optimality of the solution. Furthermore, finding the correct model parameters is challenging and requires many heuristic search processes. 

To address this challenging but practical issue, we adopted a hyperparameter optimization scheme to find the system model parameters in a data-driven approach, which is widely used in the field of machine learning. Specifically, we used the hyperband \cite{li2017hyperband} algorithm to obtain the parameters of a simplified Pacejka tire model \cite{pacejka2005tire} for vehicle dynamics. The hyperband algorithm is a variation of random search, but with some explore-exploit theory to find the best hyperparameters for each configuration based on an evaluation loss. It defines budget $B$, which is the total number of epochs required to find the hyperparameter configuration. $B$ determines the number of samples and the number of iterations spent on each sampled configuration. Comparing the evaluation loss for each sampled configuration, hyperband excludes the configuration with a large loss and allocates more budget for the configuration with a low loss. It repeats this process until the optimal configuration is chosen. To handle the local minimum problem, we apply Gaussian mutation process \cite{yu2020hyper}. New parameter configurations can be generated by injecting Gaussian noise to the selected configurations, and this increases the likelihood of finding a well-fitted solution with a lower evaluation loss.

Algorithm \ref{algo:hyper} summarizes the aforementioned parameter search process, where $R$ is the maximum budget amount and $\eta$ is an input that controls the proportion of the discarded configurations. For parameter optimization, we define the following three functions in the hyperband.

\begin{itemize}
\item \textit{$\text{get\_hyperparameter\_configuration}(n)$: } a function that returns a set of $n$ hyperparameter configurations from normal distribution defined over the configuration space. 
 
\item \textit{$\text{mutate\_then\_return\_eval\_loss}(t,r_i)$: } a function that receives a hyperparameter configuration $t$ and a resource (budget) allocation $r$ as arguments and returns the evaluation loss after mutating the configuration for the allocated resources.
 
\item \textit{$\text{select\_top\_k\_configuration}(T, L, \lfloor n_i/\eta \rfloor)$: } a function that receives a set of hyperparameter configurations $T$ and their corresponding evaluation losses $L$ and returns the top $k$ performing configurations.
 
\end{itemize}

\begin{algorithm} [t!]
\caption{Hyperband algorithm for model identification parameter search}
 
\textbf{Input: } $R, \eta$ $(\text{default } \eta = 3)$
\label{algo:hyper}
\begin{algorithmic}[1]
\State $ s_{max} \leftarrow \lfloor \text{log}_{\eta}(R) \rfloor, B = (s_{max} + 1)R $
 
\For{$s \in \{s_{max},s_{max}-1, ..., 0 \}$}
    \State $n = \lceil \frac{B}{R} \frac{\eta^{s}}{(s+1)} \rceil, r = R\eta^{-s}$
    % \Statex{// Begin SHA with $(n, r)$ inner loop}
    \State $T = \text{get\_hyperparameter\_configuration}(n)$ 
    \For{$i \in \{ 0, ..., s \}$}
    \State $ n_i = \lfloor n\eta^{-i} \rfloor $
    \State $ r_i = r\eta^i $
    % \State $ L = \{ \text{run\_then\_return\_val\_loss}(t,r_i) : t \in T \} $
    % \State $ T = \text{top\_k}(T, L, \lfloor n_i/\eta \rfloor) $
    \State $ L = \{ \text{mutate\_then\_return\_eval\_loss}(t,r_i) : t \in T \} $
    \State $ T = \text{select\_top\_k\_configuration}(T, L, \lfloor n_i/\eta \rfloor) $    
    \EndFor
\EndFor
 
\end{algorithmic}
\textbf{Output: } \text{Configuration with the smallest loss} $L$
\end{algorithm}

\section{Game-theoretic model predictive controller}

\subsection{Overall Structure}
The proposed game-theoretic MPC is designed to infer the optimal control command to drive through the track as quickly as possible without collision with surrounding opponents in a head-to-head racing scenario. The proposed controller is composed of three modules: 1) game-based opponents' trajectory predictor, 2) high-level race strategy planner, and 3) MPC-based low-level controller. The predictor establishes a two-player game between surrounding vehicles based on the perception results and solves them sequentially to set the future trajectory of the surrounding vehicles. At this time, we assumed that all the vehicle dynamics are same, and the payoff, the main gradient of the game, was shared as an objective function of the MPC. 
% Based on the predicted trajectory of opponents, the race strategy planner, which utilizes the MPC as a trajectory planner, plans the strategies according to the predefined race mode (position-keeping or overtaking).
% Based on the predicted trajectories, the racing strategy planner selects the strategies according to the predefined race mode (position-keeping or overtaking).
Based on the predicted trajectories, the racing strategy planner selects the strategies according to the predefined race mode (position-keeping or overtaking), balancing between aggressiveness and conservativeness.
Finally, the MPC-based low-level controller is responsible for generating the optimal control command along the race strategy. The overall structure of our approach is shown in Fig. \ref{fig:overview}.

\subsection{Extend Stackelberg Game to N-Player Game}
% The Stackelberg game is a simple two-player sequential game played by a leader and a follower, wherein the leader first commits his strategy, and the follower observes the leader's strategy and responds to it, while they both try to maximize their own payoff. 
The Stackelberg game is a simple two-player sequential game played by a leader and a follower. In this game, the leader first commits his strategy, and the follower observes the leader’s strategy and responds to it while both are trying to maximize their own payoff.
% The simple formulation of Stackelberg game can be represented as follows:
% {formulation}

\begin{figure}[t!]
\centering
\includegraphics[width=0.4\textwidth]{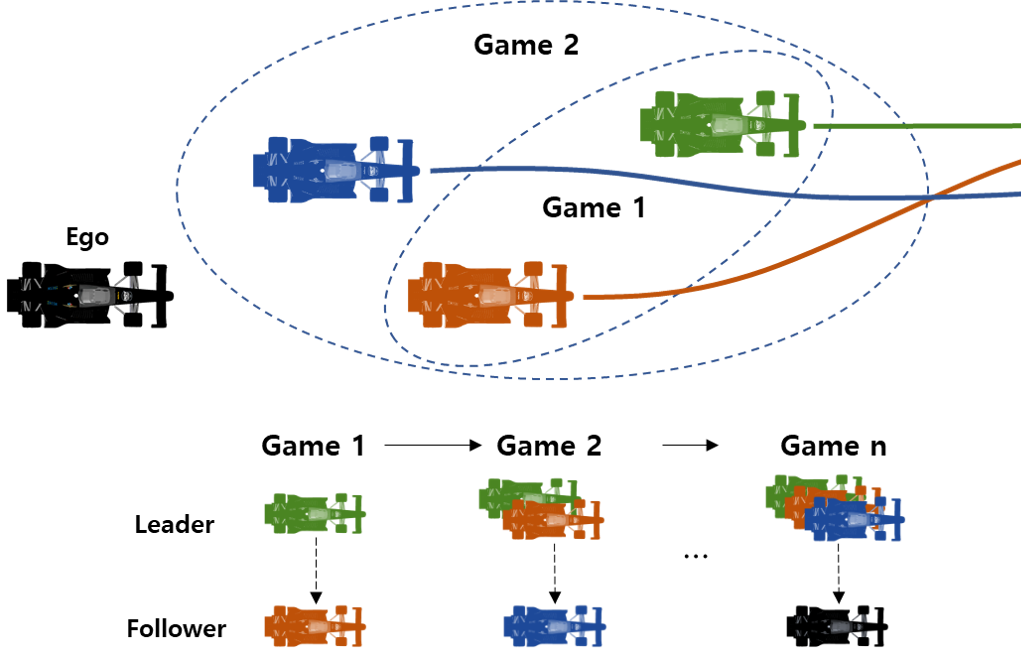}
% \caption{Structure of Stackelberg game implemented in the proposed approach.}
\caption{Visualization of the structure of the proposed Stackelberg game implementation. A two-player game (leader vs. follower) is turned into an n-player game by recursively stacking the leader's trajectories.}
\label{fig:scheme}
\end{figure}

To extend the two-player game to an n-player game, we sequentially establish independent Stackelberg games between the surrounding vehicles. Starting from the leading one, we solve each game recursively, as shown in Fig. \ref{fig:scheme}.
% To extend the two-player game to an n-player game, we sequentially established an independent Stackelberg game between the surrounding vehicles from the front leading vehicle and solved them recursively, as shown in Fig. \ref{fig:scheme}.
In each game, we assumed that two vehicles are trying to maximize their own progress, whereas only the follower takes into account the collision. Therefore, each player's payoff function can be formulated as the objective function of MPC, as described in Section \ref{sec:mpc-strategy-planner}. The output of each game (one trajectory for each player) is passed to consecutive games and set as additional leader strategies. To keep the computational load acceptable, we only considered the vehicles within a 100 m range from the ego-vehicle. 

\subsection{MPC-based Low-level Controller and High-level Race Strategy Planner} \label{sec:mpc-strategy-planner}
This section begins with the vehicle dynamics model used in this study, and the path-following MPC problem is formulated. Subsequently, it represents the detailed contents of the MPC-based low-level controller and high-level strategy planner for head-to-head autonomous racing.

\subsubsection{Vehicle Model} \label{sec:vehicle-model}
\begin{figure}[t!]
\centering
\includegraphics[width=0.37\textwidth]{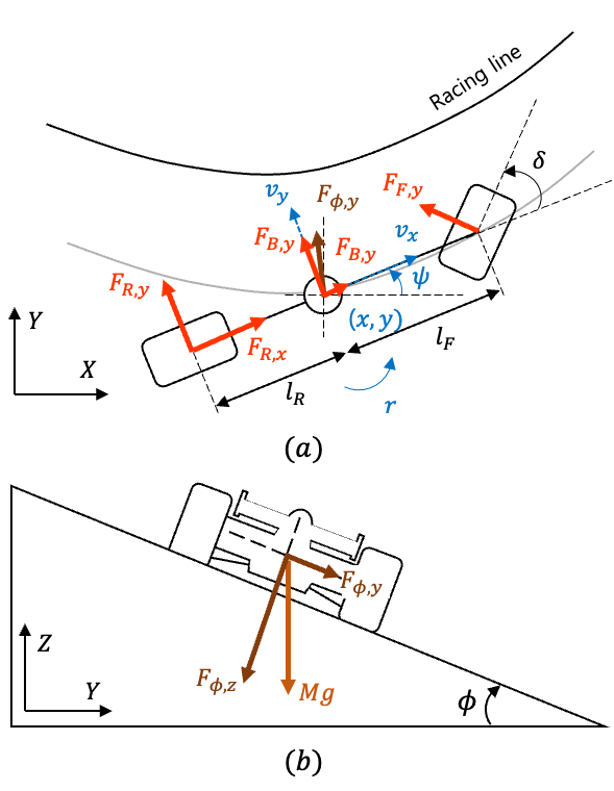}
\caption{Schematic of vehicle dynamics model (a) in the yaw and (b) roll planes.}
\label{fig:vehicle_model}
\end{figure}

The vehicle was modeled using a dynamic bicycle model \cite{kong2015kinematic}. For the sake of simplicity, the vehicle was modeled as a single rigid body expressed by mass $m$ and inertia $I_z$. In the bicycle model, lateral tire forces $F_{F,y}$ and $F_{R,y}$ were applied at the front and rear wheels, respectively. Our race vehicle was a rear-wheel drive; therefore, the longitudinal tire force $F_{R,x}$ was applied only at the rear wheel. The pitch, roll, and vertical dynamics were ignored, and only the motion on a flat surface was considered. The resulting vehicle dynamics are shown in Fig. \ref{fig:vehicle_model}, and the equation of the dynamics is given in Equation \ref{eq:vehicle_model}.

\begin{equation}
    \label{eq:vehicle_model}
    \begin{aligned}
    \begin{bmatrix}
    \dot{X} \\ \dot{Y} \\ \dot{\psi} \\ \dot{v_x} \\ \dot{v_y} \\ \dot{r} \end{bmatrix} =
    \begin{bmatrix}
    v_x \cos{\psi}-v_y \sin{\psi} \\
    v_x \sin{\psi} + v_y \sin{\psi} \\
    r \\
    \frac{1}{m}(F_{R,x}-F_{F,y}\sin{\delta} + m v_y r + F_{B,x}) \\
    \frac{1}{m}(F_{R,y} + F_{F,y}\cos{\delta}-m v_x r + F_{B,y}) \\
    \frac{1}{I_z}(F_{F,y} l_F \cos{\delta}-F_{R,y} l_R) \\
    \end{bmatrix},
    \end{aligned}
\end{equation}
where $X$, $Y$ and $\psi$ are the position and orientation in a global coordinate, respectively. $v_x$, $v_y$ and $r$ are the longitudinal, lateral velocities and yaw rate, respectively. $l_F$ and $ l_R$ are the distances from the center of gravity to the front and rear wheels, respectively. The model is represented by the 6-dimensional state $\textbf{x} = [X,Y,\psi,v_x,v_y,r]^T$ and 2-dimensional input $\textbf{u} = [\delta, D]^T$, where $\delta$ and $D$ are the steering angle and throttle command, respectively. Note that the braking command is included in the negative range of the $D$ term.

The tire forces are represented by the simplified Pacejka tire model,
\begin{equation}
    \label{eq:tire_model}
    \begin{aligned}
    \alpha_F &= \arctan{\frac{v_y + l_F r}{v_x}-\delta}, \\
    \alpha_R &= \arctan{\frac{v_y-l_R r}{v_x}}, \\
    F_{i,y} &= D_i \sin{C_i \arctan{B_i \alpha_i}}, i \in \{F,R\}, \\
    F_{R,x} &= C_m D-C_{r}-C_{d} v_x^2 \\
    \end{aligned}
\end{equation}
where $B_i, C_i, D_i, i \in \{F,R\}, C_m$ are tire parameters, and $\alpha_i, i \in \{F,R\}$ are the front and rear slip angles, respectively. The tire parameters were identified as described in Section \ref{sec:vehicle-model}, while the rolling resistance, $C_r$, was identified from the experiments. In addition, we incorporated the wake effect of the leading vehicle \cite{guerrero2020aerodynamic}. We used an adaptive drag parameter $C_d$, to exploit drafting effect.

\subsubsection{Model Predictive Contouring Control}
Motivated by \cite{ liniger2015optimization}, we adapted a MPCC scheme that accomplishes the path-following task of a vehicle in an optimal fashion via MPC. The typical MPC problem can be stated as follows:
\begin{subequations} 
\begin{align}
\min_\textbf{U} \quad
& J \label{eq:general-mpc-objective} \\
\text{s.t}\quad
& \textbf{x}_{k+1} = \textbf{f}(\textbf{x}_{k}, \textbf{u}_{k}), \quad k = 0, ..., N-1 \label{eq:general-mpc-equality} \\
& \textbf{x}_k \in \mathcal{X_C}, \textbf{u}_k \in \mathcal{U_C} \\
& \textbf{x}_N \in \mathcal{X_C}_N \\
& \textbf{x}_0 = \textbf{x}(t_0)
\end{align}
\end{subequations}
% where $J$ is the cost function, $\textbf{U} = \{ \textbf{u}_{0}, ..., \textbf{u}_{N-1} \}$ is the control sequence, which is the optimization variable, $\mathcal{X_C}$ and $\mathcal{U_C}$ are safe sets for state and control, $\mathcal{X_C}_N$ is the terminal safe set, $\textbf{x}_N $ is the terminal state, and $\textbf{x}_0$ is the state $\textbf{x}$ at the time of measurement $t_0$. The safe sets were determined based on the constraints at each time step.
where $J$ is the cost function, $\textbf{U} = \{ \textbf{u}_{0}, ..., \textbf{u}_{N-1} \}$ is the control sequence (i.e. the optimization variable), $\mathcal{X_C}$ and $\mathcal{U_C}$ are safe set for state and control, $\mathcal{X_C}_N$ is the terminal safe set, $\textbf{x}_N $ is the terminal state and $\textbf{x}_0$ is the state $\textbf{x}$ at the time of measurement $t_0$. The safe sets are decided based on the constraints at each time step.

By extending the typical MPC problem presented above, MPCC is an algorithm that enables the vehicle to follow the path by adding the contouring cost $Q$ and control quality cost $R$ into the cost function. Following the definitions and notations in \cite{ liniger2015optimization}, $Q$ and $R$ are formulated as follows:
\begin{equation} 
\label{countouring_cost}
Q(\textbf{x}_{k}) = q_c \hat{e}^2_{c,k} + q_l \hat{e}^2_{l,k}-\gamma v_{\theta, k}
\end{equation}
\begin{equation} 
\label{control_cost}
R(\textbf{u}_k, \Delta \textbf{u}_k) = \textbf{u}_k^\mathrm{T} \textbf{R}_\textbf{u} \textbf{u}_k + \Delta \textbf{u}_k^\mathrm{T} \textbf{R}_{\Delta \textbf{u}} \Delta \textbf{u}_k
\end{equation}
where $\hat{e}_{c,k}$ is the contouring error, $\hat{e}_{l,k}$ is the lag error, $v_{\theta, k}$ is the speed at time step {k} with respect to the reference path, and $q_c$, $q_l$, and $\gamma$ are their corresponding weights.

Unlike previous studies, we directly used the future trajectories of surrounding vehicles predicted through the game-based predictor as soft constraints of the MPCC. Through this, we could avoid collisions not only with static obstacles but also with various adjacent vehicles racing with their own strategies. However, in the racing scenario, because it is a non-cooperative game, errors in predicting the opponent's driving trajectory cannot be avoided. The prediction error is generated from the uncertainty in the opponent's state estimation, vehicle model, and control law. Here, we assumed that the \textit{combined uncertainty}, $\sigma$, can be approximated as a normal distribution $\mathcal{N} (0, diag(\sigma))$ following the independent and identically distributed conditions. Thus, we define the constraint in a finite horizon with a confidence interval $p_{t}$ as follows:

\begin{align} \label{collision_constraint}
%d(\mathbf{x}_{op, k}, \mathbf{x}_k)> d_{thres}(p_{t}), \quad \forall k=0,...N
d(\mathbf{x}_{op, k}, \mathbf{x}_k)> p_t \sigma, \quad \forall k=0, ..., N
\end{align}
where the distance function $d$ is the Euclidean distance from ego position $({x}_k, {y}_k)$ corresponding to $\mathbf{x}_{k}$,  to the predicted opponent position $({x}_{op, k}$, ${y}_ {op, k})$ corresponding to $\mathbf{x}_{op, k}$, and $p_t$ is the decreasing confidence interval defined for every time step.

The final MPCC formulation of our proposed controller with constraints can be rewritten as follows:
\begin{subequations}
\label{eq:mpc-all}
\begin{align}
\min_\textbf{U} \quad
& \sum_{k=0}^NQ(\textbf{x}_{k}) + R(\textbf{u}_k, \Delta \textbf{u}_k) \label{eq:final-cost} \\
\text{s.t}\quad
& \textbf{x}_{k+1} = \textbf{f}(\textbf{x}_{k}, \textbf{u}_{k}), \quad k = 0, ..., N-1 \\
& \textbf{x}_0 = \textbf{x}(t_0) \\
& \textbf{u}_{min} \leq \textbf{u}_k \leq \textbf{u}_{max} \\
& \Delta \textbf{u}_{min} \leq \Delta \textbf{u}_k \leq \Delta \textbf{u}_{max} \\
& (x_k-x_{ref,k})^2+(y_k-y_{ref,k})^2 \leq R^2_{track} \\
& d(\mathbf{x}_{op, k}, \mathbf{x}_k)> p_t \sigma
\end{align}
\end{subequations}

\subsubsection{High-Level Race Strategy Planner}
\begin{figure}[t!]
\centering
\includegraphics[width=0.44\textwidth]{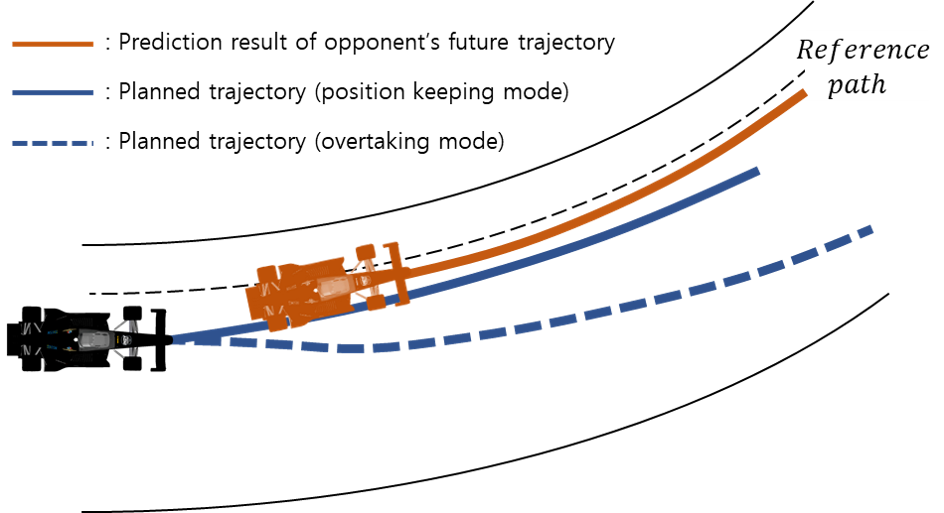}
\caption{Two different planner's outcomes in position-keeping mode (solid blue) and overtaking mode (dashed blue) based on the same trajectory prediction (solid orange).}
\label{fig:highlevel}
\end{figure}
Professional human drivers have the ability to respond to various race circumstances while balancing their driving aggressiveness and conservativeness. To implement these racing behaviors in our controller, we hierarchically added a high-level race strategy planner before the MPCC-based low-level controller. The race strategy planner is responsible for planning the race-oriented strategies of the ego-vehicle using various configurations of the MPC controller. We used the MPC solution also in the trajectory planner implementation, with the objective function equal to \eqref{eq:final-cost}. We designed two distinct modes within the planner: position-keeping and overtaking. In the position-keeping mode, the contouring cost weight $q_c$ was set to large values so that the ego-vehicle tightly follows the reference path while maximizing the aero-drafting advantage at the rear of the preceding vehicle. By contrast, the overtaking mode was configured such that the vehicle’s progress is maximized by setting the speed cost weight, $\gamma$, to a large value. The two modes are selectively utilized, and the criterion for selecting the overtaking mode is whether the progress in the terminal state exceeds a certain threshold (here, we set the threshold as 3 m) compared to the predicted progress of the competitor. Fig \ref{fig:highlevel} shows the difference in strategy between the two modes.

\section{Experiments}
\subsection{Experimental Design}

\begin{figure}[t!]
\centering
\includegraphics[width=0.45\textwidth]{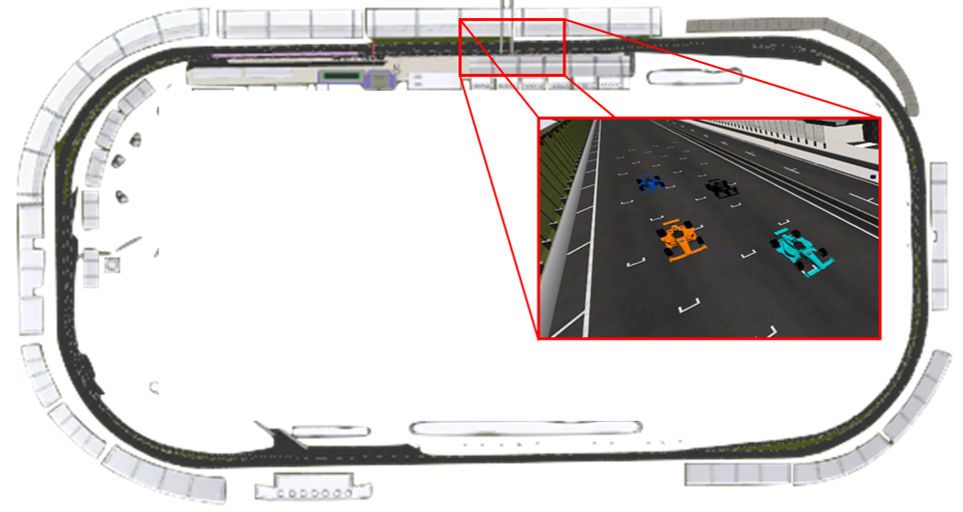}
\caption{Top view of the Indianapolis Motor Speedway (IMS) modeled in the Ansys VRXPERIENCE driving simulator.}
\label{fig:sim}
\end{figure}
We used the Ansys VRXPERIENCE driving simulator, the official software of the Indy Autonomous Challenge, as shown in Fig. \ref{fig:sim}. 
% We created a head-to-head racing scenario including four identical vehicles with a rolling start (from 100 km/h), and the ego-vehicle started from the fourth grid.
We created a head-to-head racing scenario with four identical vehicles in a rolling start setting and the ego-vehicle placed last in the group.
Here, we set the planning horizon to 1 s, and the top speed of the vehicle was limited to 300 km/h. The position and speed of the surrounding vehicles were detected by a radar provided by the simulator. To make the experiments more realistic, we added random noise to the perception result, intentionally introducing error in the predicted trajectories.

To verify our proposed approach, we conducted the following case studies:
\begin{itemize}
  \item Case 1) With and without the proposed high-level race strategy planner: This study was conducted in the curved segment of the track.
  \item Case 2) Replace the game-based predictor with the extended Kalman filter (EKF) method: This study was conducted in the straight segment of the track. For the EKF method, we assumed that the acceleration and heading did not change during the prediction. 
\end{itemize}

\subsection{Simulation Results}
% 그림 \cite{fig:result_race2}은 case 1에 대한 실험 결과를 보여준다. The first row of Fig. ?? 는 curved segment 트랙의 환경에서 ego-vehicle과 preceding vehilce의 주행 궤적을 시간에 따라 보여준다. 제안하는 race strategy planner가 적용되지 않은 경우(left)에는, MPC based low-level controller로 부터 곧바로 차량의 제어값을 계산되었다. In this situation, 우리의 차는 track boundary와의 공간이 있는 우측으로 전방 차량 추월을 시도하였다 at t=??. 하지만 차량의 동력학적인 constraint을 만족하기 위해 throttle을 release하고 이에 따라 차량의 속도는 감소하게 되며, 전방 차량은 일정한 가속도로 주행을 수행하였다. 늘어난 주행 거리와 줄어든 속도로 인해, 우리의 차량과 전방 차량과의 거리는 추월을 시작한 시점에 비해 늘어난 것을 확인할 수 있다. 이후 우리 차량은 다시 contouring error를 줄이기 위해 reference path로 복귀하기 위한 제어를 수행하였다. 반면, 제안하는 race strategy planner를 사용한 경우는 (right) 전방 차량의 후미에서 주행였다. 이때 차량은 aero drafting advantage를 통해 더 가속할 수 있지만 차량을 추월하기 충분하지 않다고 판단하고 throttle release motion을 통해 전방 차량과 근접한 거리를 유지하며 주행하였다. 비록 두 경우 모두 전방 차량을 추월하지는 못하였지만, t=??에서의 주행 속도 차이는 약 ?? km/h로 극명한 차이를 보였다. 

\begin{figure}[t!]
% \centering
\includegraphics[width=0.5\textwidth]{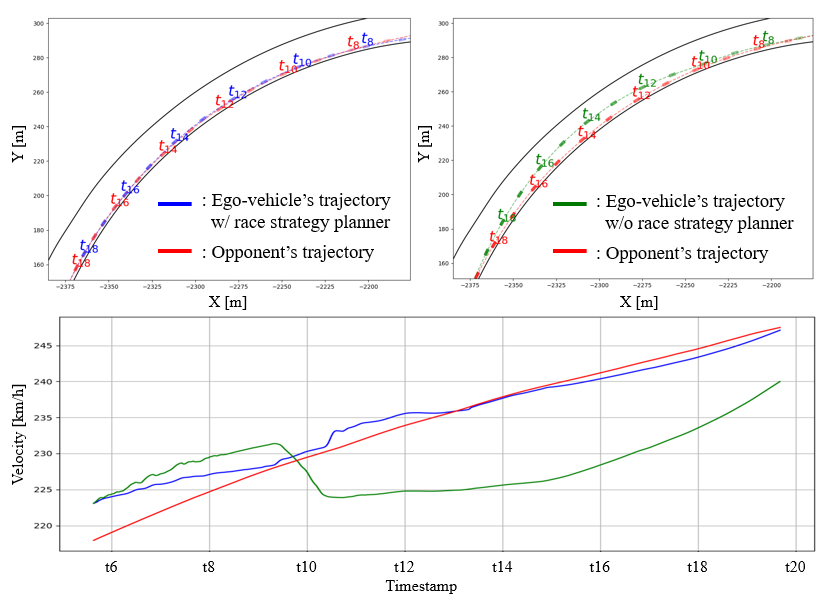}
\caption{Simulation results in case 1.}
\label{fig:result_race2}
\end{figure}

Fig. \ref{fig:result_race2} shows the experimental results of case 1. The first row of Fig. \ref{fig:result_race2} shows the trajectories of the ego and the preceding vehicles over time in a curved segment of the track. 
% In the case of the proposed race strategy planner not being applied (first row, right), the control commands were directly calculated from the MPC based low-level controller.
Without the race strategy planner (first row, right), the control commands were directly calculated from the MPC based low-level controller.
Here, the ego-vehicle attempted to overtake the opponent to the right side at t=10. However, the throttle was released to satisfy the vehicle dynamics constraints and, accordingly, the vehicle speed decreased while the preceding vehicle ran at constant acceleration. Because of the increased travel distance and reduced speed, the distance gap between the ego and the opponent increased. After that, the ego-vehicle returned to the reference path to reduce contouring error and maximize its progress. 
% On the other hand, when the proposed race strategy planner was applied (first row, left), it was driven in position-keeping mode in the planner.
On the other hand, with the race strategy planner (first row, left), the ego-vehicle was driven in position-keeping mode by the planner.
In this case, the ego-vehicle could accelerate more using the aero drafting advantage, but the acceleration was not enough to overtake the other. 
% Therefore, ego-vehicle drove while maintaining a close distance to the vehicle ahead through the throttle releasing motion. 
Therefore, the ego-vehicle kept the reference path driving close to the opponent and releasing the throttle, where necessary. Although there was no overtaking in both cases, there was a huge difference in terms of driving speed, as shown in the second row of Fig. \ref{fig:result_race2}.

% Next, 그림 ??은 제안하는 game-based predictor의 효과를 보여주기 위한 case 2에 대한 실험 결과를 보여준다. 우리는 다양한 opponents와의 interaction이 다수 존재하는 race 초반에서 실험을 수행하였다. To demonstrate the overtaking, 우리는 opponents' initial speed and ours were set to 100 km/h and 110 km/h, respectively. The first and second row of the Fig. ??는 제안하는 game-based predictor를 통해 주변 차량의 미래 궤적을 예측하고 이를 통해 주행한 결과를 시간에 따라 보여준다. 우리의 차량은 t=5??까지는 자신의 reference path와의 contouring error를 줄이기 위해 트랙의 우측으로 주행하는 모습을 볼 수 있다. 이때 특정 순간부터 주변 차량들간의 interaction을 예측하고 우리의 차량은 주변 차량과의 충돌이 존재하지 않는 좌측으로의 주행한다. 반면 주변 차량간의 interaction을 반영하지 못하는 EKF method를 바탕으로 주변 차량의 궤적을 예측하는 경우(third and fourth rows of the Fig. ??)에는, t=??에서 초록 차량과 충돌이 발생하였다.

\begin{figure*}[t]
    \centering
    \includegraphics[width=0.86\textwidth]{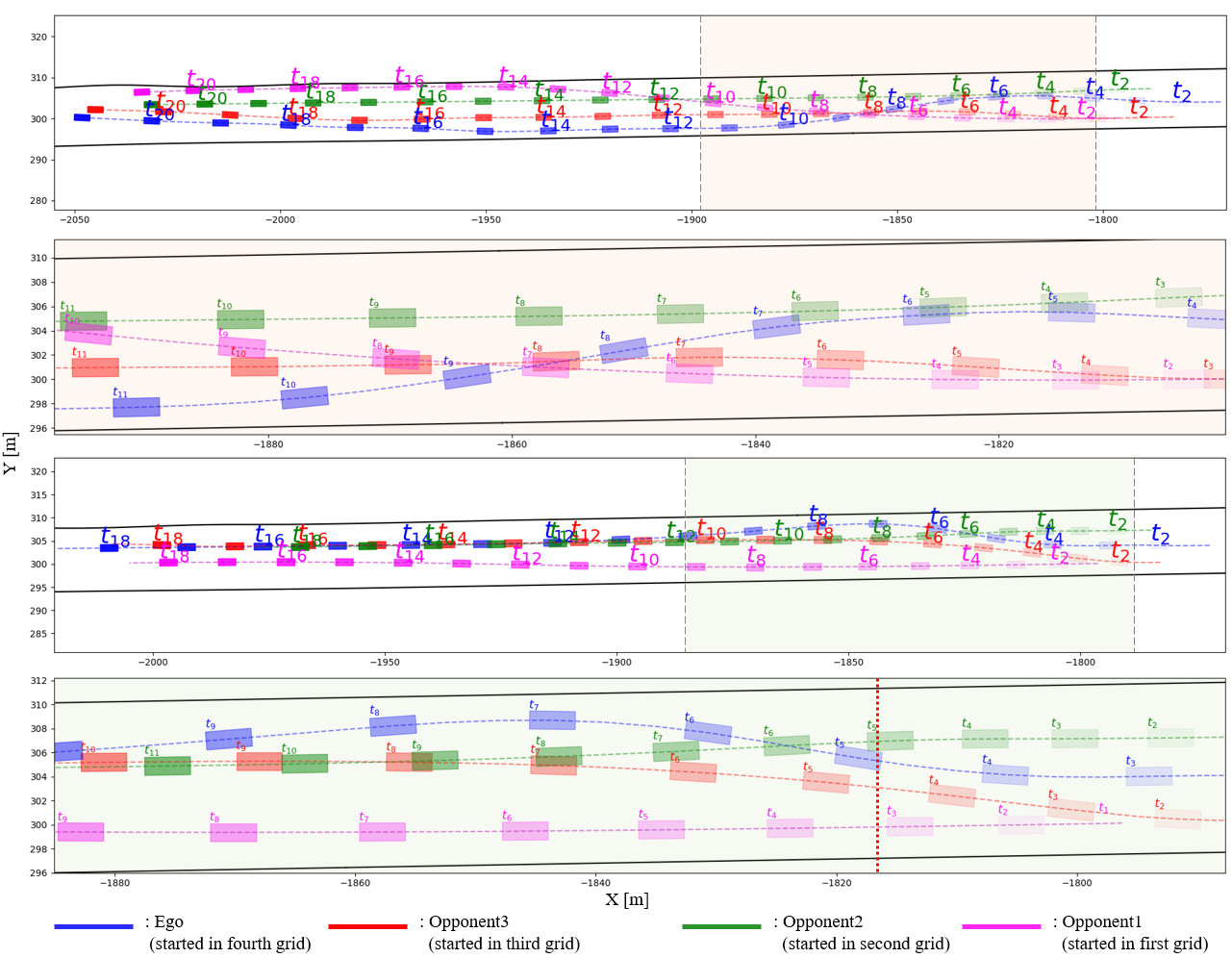}
    \caption{Simulation results in case 2.}
    \label{fig:result_race1}
\end{figure*}

Next, Fig. \ref{fig:result_race1} presents the experimental results for case 2, which show the effectiveness of the proposed game-based predictor. We conducted experiments at the beginning of the race, where there are many interactions with competitors. Also, opponents' initial speed and ours were set to 100 km/h and 110 km/h, respectively. The first and the second rows of Fig. \ref{fig:result_race1} show the results of the proposed predictor in action.
% when the ego-vehicle predicts the future trajectory of others through the proposed game-based predictor. 
Here, the ego-vehicle drove to the right side of the track until t=5 to reduce the contouring error with its reference path. 
After t=5, the ego-vehicle reacted to the predicted competitors' trajectories and drove to the left side of the track, aiming to avoid the collision without braking maneuver.
On the other hand, when the EKF was applied as a predictor (third and fourth rows of Fig. \ref{fig:result_race1}), a collision occurred around t=5. This result clearly shows the effectiveness of our approach, that has the advantage of predicting the opponents' future trajectories based on the interaction between them.

% 이 결과는 레이싱에서 주변 차량 간의 interaction을 바탕으로 주행 궤적을 예측하는 우리의 approach의 effectiveness를 clear하게 보여준다. 

% EKF를 주변 차량의 궤적 predictor로 사용했을 경우에는 high-level race strategy planner 유무와 관계없이 모두 충돌이 발생하여 lap time은 do not finish(DNF)로 측정되었다. 반면 제안하는 game-based opponents’trajectory predictor를 사용하였을 경우에는, 충돌이 발생하지 않았다. 나아가서 이와 함께 high-level race strategy planner를 사용하였을 경우, 사용하지 않았을때에 비해 약 1.1초의 lap time 감소를 시킬 수 있었다. 

\begin{table}[t!]
\caption{Single lap simulation results.}
\begin{tabular}{cc|ccc}
\hline
\begin{tabular}[c]{@{}c@{}}Predictor\\ type\end{tabular} & \begin{tabular}[c]{@{}c@{}}High-level race \\ strategy planner\end{tabular} & \begin{tabular}[c]{@{}c@{}}Collision\\ occurred\\ time (sec)\end{tabular} & \begin{tabular}[c]{@{}c@{}}\# of \\ overtaking\end{tabular} & \begin{tabular}[c]{@{}c@{}}Single \\ lap time\\ (sec)\end{tabular} \\ \hline
EKF & W/O & t = 7.3 & - & DNF \\ \hline
EKF & W/  & t = 8.8 & - & DNF \\ \hline
\begin{tabular}[c]{@{}c@{}}Game\\ based\end{tabular} & W/O & - & 0 & 53.135 \\ \hline
\begin{tabular}[c]{@{}c@{}}Game\\ based\end{tabular} & W/  & - & 1 & 52.051 \\ \hline
\end{tabular}
\label{tab:result}
\end{table}

Single lap simulation results are listed in Table \ref{tab:result}. When EKF was used as trajectory predictor, collisions occurred regardless of the presence or absence of a high-level race strategy planner, and the lap time was measured as do not finish (DNF). On the other hand, when the proposed game-based opponents’ trajectory predictor was used, collision had not occurred. Furthermore, when a high-level race strategy planner was used with this predictor, 1.1 sec of lap time gain was observed.

\section{CONCLUSIONS}
In this study, we introduced a game-theoretic MPC for head-to-head autonomous racing and a data-driven approach for model identification. 
% For model identification, we explored tire model parameters with high nonlinearity from the driving data by applying the hyperband algorithm widely used in the machine learning field.
% For the latter, we explored the highly non-linear tire model by applying the Hyperband algorithm, widely used in the machine learning field.
For model identification, we explored the parameters of the highly non-linear tire model from the driving data by applying the hyperband algorithm, widely used in the machine learning field.
% For model identification, 우리는 뉴럴 모델 학습에 사용되는 hyperband algorithm을 적용하여 MPC의 바탕이 되는 차량 모델 파라미터를 구하였다. 제안하는 game-theoretic MPC는 주변 차량 간의 interaction이 반영된 미래 궤적 predictor, 다양한 race 상황에 대응하기 위한 전략 planner, 그리고 vehicle dynamics를 고려한 MPC-based low-level controller로 구성되어있다. 제안하는 approach의 성능은 IAC 공식 시뮬레이터에서 검증되었다. 실험 결과는 제안하는 각 모듈을 통해 다양한 레이스 상황에 적절하게 대응하면서도 최소 lap time을 보여주는 것을 보여준다. 
The proposed game-theoretic MPC comprises three modules: 1) Stackelberg game-based opponents' trajectory predictor, 2) high-level race strategy planner, and 3) MPC-based low-level controller. The performance of the proposed method was demonstrated based on various scenarios in a highly realistic racing simulator. 
% Experimental results show that the proposed method can be deployed on a track with the fastest lap time without collision in a head-to-head race. 
% 실험 결과는 제안하는 approach를 통해 head-to-head racing 상황에서 충돌을 회피하면서 빠른 속도로 주행하는 것을 보여주었다. 우리는 이번 연구 결과를 확장하여, 전략적 blocking과 deep learning 기반의 race mode 선택에 대해 연구할 계획이다. 
The experimental results showed that the proposed approach drives as quickly as possible while avoiding collision in a head-to-head racing situation. 

We plan to expand the results of this study by adding a strategic blocking mode in the high-level race strategy planner and reinforcement learning based head-to-head autonomous driving.

% We would like to further solve the practical issues of real-world head-to-head racing problems running at near-maximum performance or where the racing performance gaps between vehicles are not significant.

\addtolength{\textheight}{-12cm}   % This command serves to balance the column lengths

%%%%%%%%%%%%%%%%%%%%%%%%%%%%%%%%%%%%%%%%%%%%%%%%%%%%%%%%%%%%%%%%%%%%%%%%%%%%%%%%

%%%%%%%%%%%%%%%%%%%%%%%%%%%%%%%%%%%%%%%%%%%%%%%%%%%%%%%%%%%%%%%%%%%%%%%%%%%%%%%%



% \section*{ACKNOWLEDGMENT}

% The preferred spelling of the word ÒacknowledgmentÓ in America is without an ÒeÓ after the ÒgÓ. Avoid the stilted expression, ÒOne of us (R. B. G.) thanks . . .Ó  Instead, try ÒR. B. G. thanksÓ. Put sponsor acknowledgments in the unnumbered footnote on the first page.

%%%%%%%%%%%%%%%%%%%%%%%%%%%%%%%%%%%%%%%%%%%%%%%%%%%%%%%%%%%%%%%%%%%%%%%%%%%%%%%%

% References are important to the reader; therefore, each citation must be complete and correct. If at all possible, references should be commonly available publications.

\bibliographystyle{unsrt}
\bibliography{reference}

\begin{thebibliography}{10}

\bibitem{roborace}
Roborace.
\newblock \url{https://roborace.com/}.
\newblock Accessed: 2021-04-20.

\bibitem{iac}
Indy autonomous challenge.
\newblock \url{https://www.indyautonomouschallenge.com/}.
\newblock Accessed: 2021-04-20.

\bibitem{racer}
Darpa’s robotic autonomy in complex environments with resiliency.
\newblock \url{https://www.darpa.mil/news-events/2020-10-07}.
\newblock Accessed: 2021-04-20.

\bibitem{kabzan2020amz}
Juraj Kabzan, Miguel~I Valls, Victor~JF Reijgwart, Hubertus~FC Hendrikx, Claas
  Ehmke, Manish Prajapat, Andreas B{\"u}hler, Nikhil Gosala, Mehak Gupta, Ramya
  Sivanesan, et~al.
\newblock Amz driverless: The full autonomous racing system.
\newblock {\em Journal of Field Robotics}, 37(7):1267--1294, 2020.

\bibitem{liniger2015optimization}
Alexander Liniger, Alexander Domahidi, and Manfred Morari.
\newblock Optimization-based autonomous racing of 1: 43 scale rc cars.
\newblock {\em Optimal Control Applications and Methods}, 36(5):628--647, 2015.

\bibitem{buyval2017deriving}
Alexander Buyval, Aidar Gabdulin, Ruslan Mustafin, and Ilya Shimchik.
\newblock Deriving overtaking strategy from nonlinear model predictive control
  for a race car.
\newblock In {\em 2017 IEEE/RSJ International Conference on Intelligent Robots
  and Systems (IROS)}, pages 2623--2628. IEEE, 2017.

\bibitem{wang2019game1}
Mingyu Wang, Zijian Wang, John Talbot, J~Christian Gerdes, and Mac Schwager.
\newblock Game theoretic planning for self-driving cars in competitive
  scenarios.
\newblock In {\em Robotics: Science and Systems}, 2019.

\bibitem{liniger2019noncooperative}
Alexander Liniger and John Lygeros.
\newblock A noncooperative game approach to autonomous racing.
\newblock {\em IEEE Transactions on Control Systems Technology},
  28(3):884--897, 2019.

\bibitem{li2017hyperband}
Lisha Li, Kevin Jamieson, Giulia DeSalvo, Afshin Rostamizadeh, and Ameet
  Talwalkar.
\newblock Hyperband: A novel bandit-based approach to hyperparameter
  optimization.
\newblock {\em The Journal of Machine Learning Research}, 18(1):6765--6816,
  2017.

\bibitem{pacejka2005tire}
Hans Pacejka.
\newblock {\em Tire and vehicle dynamics}.
\newblock Elsevier, 2005.

\bibitem{yu2020hyper}
Tong Yu and Hong Zhu.
\newblock Hyper-parameter optimization: A review of algorithms and
  applications.
\newblock {\em arXiv preprint arXiv:2003.05689}, 2020.

\bibitem{kong2015kinematic}
Jason Kong, Mark Pfeiffer, Georg Schildbach, and Francesco Borrelli.
\newblock Kinematic and dynamic vehicle models for autonomous driving control
  design.
\newblock In {\em 2015 IEEE Intelligent Vehicles Symposium (IV)}, pages
  1094--1099. IEEE, 2015.

\bibitem{guerrero2020aerodynamic}
Alex Guerrero and Robert Castilla.
\newblock Aerodynamic study of the wake effects on a formula 1 car.
\newblock {\em Energies}, 13(19):5183, 2020.

\end{thebibliography}

\end{document}